# Segmentation overlapping wear particles with few labelled data and imbalance sample

Peng Peng*, Jiugen Wang
Faculty of Mechanical Engineering, Zhejiang University, Hangzhou, 310027, China
*Corresponding author: Peng Peng
E-mail: pengpzju@163.com

**Abstract**:
Ferrograph image segmentation is of significance for obtaining features of wear particles. However, wear particles are usually overlapped in the form of debris chains, which makes challenges to segment wear debris. An overlapping wear particle segmentation network (OWPSNet) is proposed in this study to segment the overlapped debris chains. The proposed deep learning model includes three parts: a region segmentation network, an edge detection network and a feature refine module. The region segmentation network is an improved U shape network, and it is applied to separate the wear debris form background of ferrograph image. The edge detection network is used to detect the edges of wear particles. Then, the feature refine module combines low-level features and high-level semantic features to obtain the final results. In order to solve the problem of sample imbalance, we proposed a square dice loss function to optimize the model. Finally, extensive experiments have been carried out on a ferrograph image dataset. Results show that the proposed model is capable of separating overlapping wear particles. Moreover, the proposed square dice loss function can improve the segmentation results, especially for the segmentation results of wear particle edge.
**Keywords**: ferrograph images; overlapping wear particles; square dice loss; edge detection; wear particle segmentation

## 1 Introduction

Ferrograph is an established technology for condition monitoring of mechanical equipment. It presents wear particles produced by mechanical equipment through ferrograph images [1]. By analyzing the wear particles in ferrograph images, researchers can determine the wear status of the mechanical equipment. For example, by analyzing the types of wear particles in ferrograph images, the wear mechanism and wear source of equipment can be recognized [2]. By analyzing the number and the size of wear particles in ferrograph image, the wear trend of mechanical equipment is obtained [3, 4].

In the early of ferrograph technology, ferrograph image analysis is highly depended on expert experience. However, manual analysis is time-consuming and a subjective result is often obtained. Hence, researches proposed ferrograph image processing methods to analyze ferrograph images automatically [5, 6]. Among the ferrograph image processing technologies, ferrograph image segmentation is a vital step for extracting wear particle features. Once the ferrograph images are segmented, the shape, edge and area of wear particles can be extracted. These wear particle features are not only helpful to determine the wear mechanism of the equipment but also to analyze the wear state of the equipment [7]. However, the wear particles in ferrography images often stick together, which makes the task of automatic segmentation of wear particles very challenging. Yu et al. proposed background colour recognition method to segment debris and background [8], Wang et al proposed the K-means clustering segmentation method for separating wear particles from the background [9]. However, these segmentation methods only realize the separation of the debris and the background, while the original overlapping particles still maintain the bond relationship. Therefore, it is difficult to count the number of wear particles in ferrograph images and obtain reliable information reflecting the wear rate of mechanical equipment. In order to segment overlapping debris, Wang et al. combined an ant colony algorithm and watershed algorithm to achieve the segmentation of overlapping debris [10]. After that, they combined the grey correlation analysis method to accelerate the segmentation of overlapping wear particles [11]. Wu et al. used mathematical morphology method to realize the segmentation of overlapping wear particles [12]. Considering that watershed segmentation is easy to lead to over-segmentation, Peng et al proposed two region merging rules to improve the segmentation results of the watershed algorithm [13]. However, the above-mentioned overlapping debris segmentation method needs to extract the features of wear particles artificially, and combine these artificially designed features to achieve better segmentation results.

In recent years, deep learning has been proved to be able to learn the features of objects in images, and it has been widely used in natural image segmentation and medical image segmentation [14-16]. In the field of ferrograph image processing, the deep learning method has also been well verified [17]. Peng et al. proposed a transfer learning method to identify overlapping wear particles in ferrograph images [18]. Peng et al. combined convolution neural network and support vector machine to realize the recognition of multiple wear particles [19]. Peng et al. proposed a wear particle detection method for ferrograph image [20]. Wang et al. designed an optimized convolution network model to identify sliding and fatigue wear particles [21]. Zhang proposed a metric learning method to solve the classification of unknown types of wear particles [22]. Although a large number of deep learning methods have been verified in ferrograph image processing, it is rarely involved in ferrograph image segmentation, especially the segmentation of overlapping wear particles. The main reason is that deep learning method has two difficulties in the field of ferrograph image segmentation: (1) deep learning method needs a lot of training samples. However, the workload of data acquisition and labelling for ferrograph image segmentation is very large; (2) the phenomenon of wear debris overlap is serious, and thus the wear debris separation task is very challenging. In the field of computer vision, researchers have also proposed few shot segmentation algorithms. Ronneberger proposed a network called U-Net for medical image segmentation with few samples. Their experiments show that only 30 labelled medical images can obtain nice segmentation results. Zhao et al. proposed an automatic data enhancement method for MRI image segmentation with few labelled data [23]. Shaban et al. proposed a dual branch network for image segmentation with few samples [25]. Wang et al. proposed a prototype network method for image segmentation [26]. Zhang et al. proposed a dense comparison module and iterative optimization module for small sample image segmentation [27]. For the segmentation of overlapping objects, Ronneberger proposed to use the weighted cross-entropy loss function to segment overlapping cells [23]. Besides, instance segmentation algorithms can be used to segment overlapped objects, such as two-stage Mask-RCNN network [28] and one-stage network blendmask [29]. However, the above research is mainly to solve the segmentation problem in medical images or natural images, and the segmentation of ferrograph images with few samples and overlapping wear particles has not been studied.

Considering the problems of few samples and overlapping of wear particles in ferrograph image segmentation, a new algorithm of overlapping debris segmentation is proposed in this paper. Firstly, an improved U-Net network is used to realize the segmentation of wear particles and background, and then the wear particle edge detection network is introduced to obtain the edge of each wear particle. Finally, a feature enhancement module is designed to fuse the low-level details and high-level semantic information, and output the final segmentation results. Considering the serious sample imbalance problem in edge segmentation, we propose a square dice loss function to optimize the model.

## 2 Method

Ronneberger proposed the U-Net model shown in Figure 1 to solve medical image segmentation [23]. The U-Net model adopts encoding and decoding network structure. And introduce skip connection to fuse low-level detailed information and high-level semantic information. In order to solve the problem of few training samples, a method of elastic deformation data enhancement is proposed. At the same time, a weighted cross-entropy loss function is proposed to separate overlapping cells. The experimental results show that only 30 medical images are used to obtain a good image segmentation effect. Inspired by their work, we propose a network model as shown in Figure 2 to solve the segmentation of overlapping wear particles with few samples.

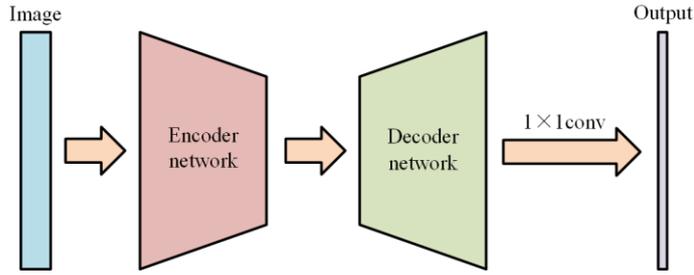

Figure 1. The framework of the U-Net model

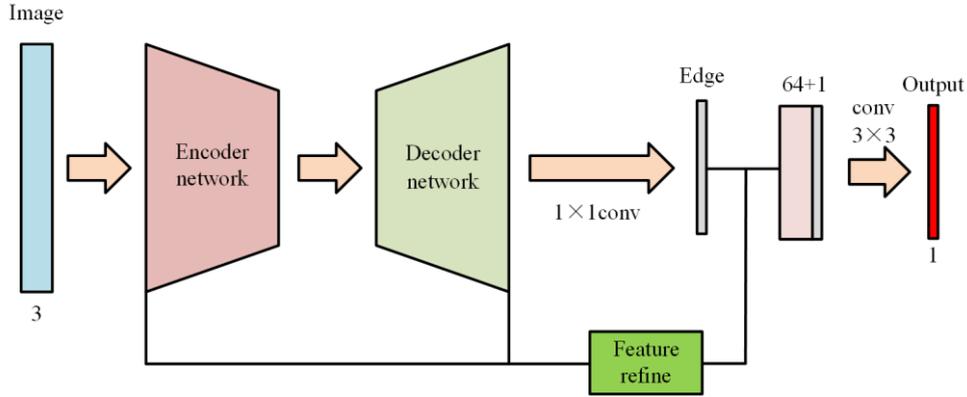

Figure 2. The structure of the proposed OWPSNet model

As shown in Figure 2, the model proposed in this study includes the following parts: an improved U-Net network, which adds a normalization operation to the original U-Net network; a wear particle edge detection network, which is mainly responsible for extracting the edge of wear particles; the feature refine module is mainly used to enhance the feature for wear particles segmentation. At the same time, considering the sample imbalance phenomenon, a new loss function, the square dice loss function, is proposed.

In order to verify the influence of the feature refine module on the network, this paper also proposes the following network that does not include the feature refine module.

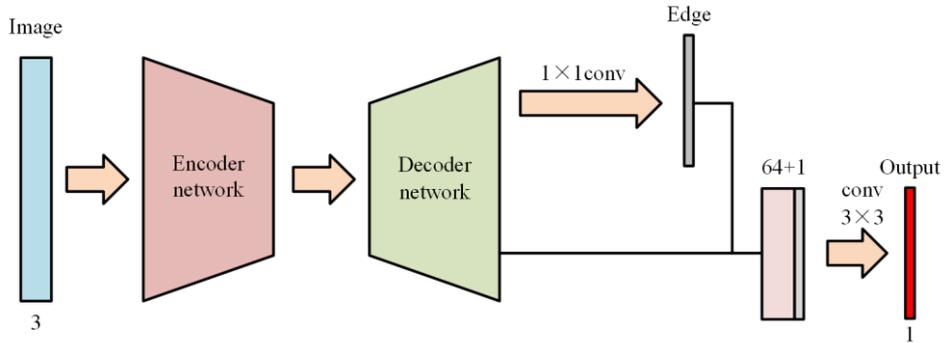

Figure 3. The proposed OWPSNet model without feature refine module

## 2.1 Wear particle edge detection

Reference [30, 31] proposed deep convolutional neural network for semantic edge segmentation. The semantic edge detection algorithm realizes image segmentation by detecting the edge of the target. But when two overlapping targets have the same semantics, the semantic edge detection algorithm cannot separate them.

In order to realize the separation of overlapping wear particles, this paper designs an edge detection branch network as shown in Figure 2. This branch network is parallel to the area segmentation network Thus, the two networks can share parameters and reduce network overfitting.

## 2.2 Feature refine module

The wear particles in the ferrograph image overlap seriously, and accurately obtaining the edges of the wear particles puts forward higher requirements on the feature extraction ability of the network. For predicting the edges of wear particles, the low-level feature maps can provide more detailed information, while the high-level feature maps can obtain more semantic information. In order to accurately obtain the edges of the abrasive grains, the two parts of information need to be well integrated. To this end, this paper proposes a feature refine module as shown in Figure 3. The first-layer feature map of the encoder convolution in U-Net is concatenated with the last-layer feature map of the decoder. Finally, the spatial attention mechanism and channel attention mechanism network proposed in [32] are introduced to obtain the final refined characteristics.

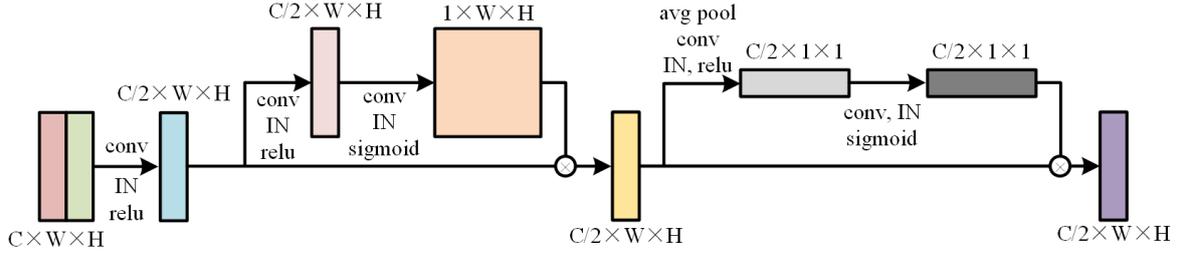

Figure 4. The feature refine module

## 2.3 Normalization

BN (Batch normalization) [33] has become a standard module in deep convolutional networks. However, research shows that the batch size of model training must be large enough to make BN obtain better results. When the batch is small, the effect of BN operation will be greatly discounted [34, 35]. Generally, image segmentation tasks have higher requirements for GPU, and only a small batch can be trained on a single GPU. In addition, the background colour of the ferrograph image and the illumination of the image are varied, and how to extract the characteristics of wear particles from the changing background also need the model to pay attention to. Research shows that IN (Instance normalization) [36] can learn the content information of the image from images in different styles and filter to the information reflecting the appearance of the image [37]. In order to design a good normalization module, this paper will add four normalization modules to the U-Net network. Figure 5 shows the four normalization modules. In this paper, OWSNet using the normalization modules shown in Figure 5(a)-(d) will be called OWSNet-BN, OWSNet-IN, OWSNet-IN-BN, OWSNet-BN-IN.

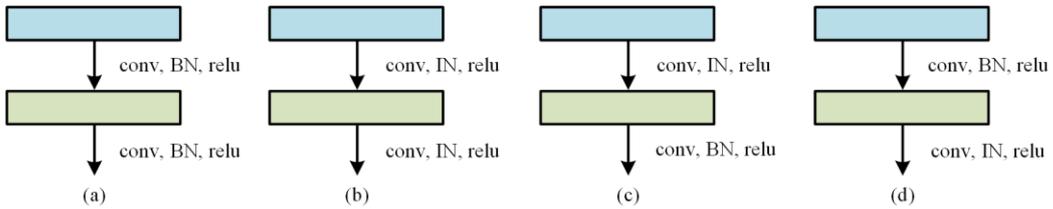

Figure 5. The four normalization modules

## 2.4 Square dice loss function

The proposed edge detection branch network is used to detect the edge of wear particles. However, the number of pixels on the edge of wear debris is far less than that of the background of ferrography image. Thus, there will be a serious sample imbalance in this study. The classical cross-entropy loss function weight all pixels equally. Therefore, the classical cross-entropy loss function is not suitable to solve the problem of sample imbalance. In order to solve the problem of unbalanced sample segmentation, Milletari et al. proposed dice loss function [38]. The formula is as follows:

$$L_{dice} = 1 - \frac{2\sum_{i}^{N} p_i t_i}{\sum_{i}^{N} p_i^2 + \sum_{i}^{N} t_i^2} \tag{1}$$

Where $p_i$ is the probability of the $i$-th pixel predicted by the network, $t_i$ is the label of the $i$-th pixel, and $N$ is the total number of pixels.

The derivative of the dice loss function to a pixel $j$ is as follows:

$$\frac{\partial L_{dice}}{\partial p_j} = -\frac{2t_j(\sum_{i}^{N} p_i^2 + \sum_{i}^{N} t_i^2) - 4p_j \sum_{i}^{N} p_i t_i}{(\sum_{i}^{N} p_i^2 + \sum_{i}^{N} t_i^2)^2} \tag{2}$$

The dice loss function is a loss function to measure regional overlap. It is different from the cross-entropy loss function. Dice loss of a pixel is not only related to the prediction probability of the network but also related to the prediction value of other pixels. From equation 2, it can be found that the denominator term of the derivative of dice loss function contains the fourth power term of network prediction probability $p$ and label $t$. therefore, when the target is small, the derivative of the function will have a large gradient, leading to the instability of training.

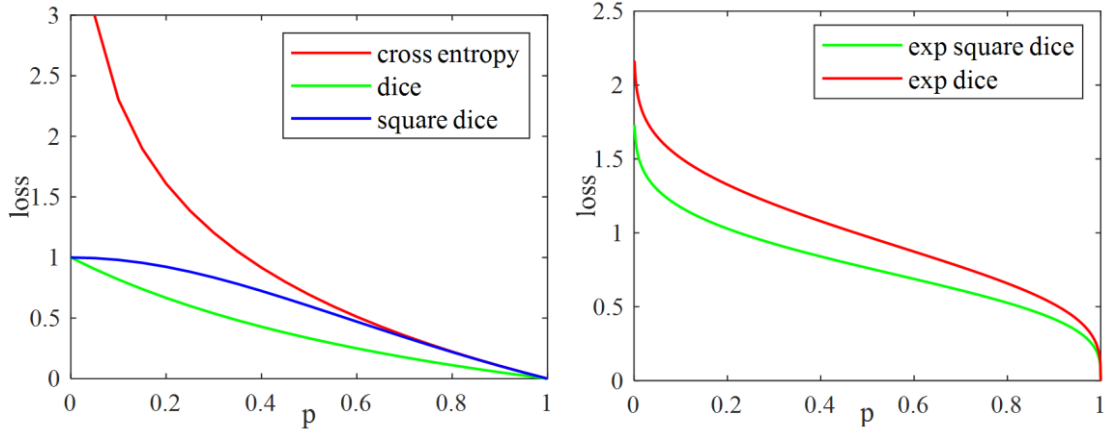

Figure 6. Different loss function curves

Considering the above shortcomings of the dice loss function, this paper proposes the following square loss function:

$$L_{square\ dice} = 1 - \frac{2\sum_{i}^{N} (p_i t_i)^2}{\sum_{i}^{N} p_i^2 + \sum_{i}^{N} t_i^2} \tag{3}$$

$$\frac{\partial L_{square\ dice}}{\partial p_j} = -4 \frac{p_j t_j^2 (\sum_{i}^{N} p_i^2 + \sum_{i}^{N} t_i^2) - p_j \sum_{i}^{N} (p_i t_i)^2}{(\sum_{i}^{N} p_i^2 + \sum_{i}^{N} t_i^2)^2} \tag{4}$$

In reference [39], an exponential logarithmic loss function is proposed to solve the problem of highly unbalanced data classification. Combined with the loss function proposed in reference [39], we propose another loss function, namely exponential square dice loss as follows:

$$L_{exp\ square\ dice} = [-\log(\frac{2\sum_{i}^{N} (p_i t_i)^2}{\sum_{i}^{N} p_i^2 + \sum_{i}^{N} t_i^2})]^{\gamma} \tag{5}$$

Figure 6 shows the curve of each loss function when there is only a single pixel. The final loss function of the model is as follows:

$$L = L_{region} + L_{edge} \qquad (6)$$

Where $L_{region}$ and $L_{edge}$ refer to the loss function of regional and edge networks respectively.

*2.5 Morphology post-processing*

In order to count the wear particles and extract the characteristic parameters of wear particles, the separation of overlapping particles is inevitable. The previous region segmentation network and edge detection network extract the region and edge of wear particles, but there may be overlapped debris not separated, as shown in the red arrow in Fig. 6(e). Therefore, it is necessary to further improve the segmentation results of overlapping wear particles.

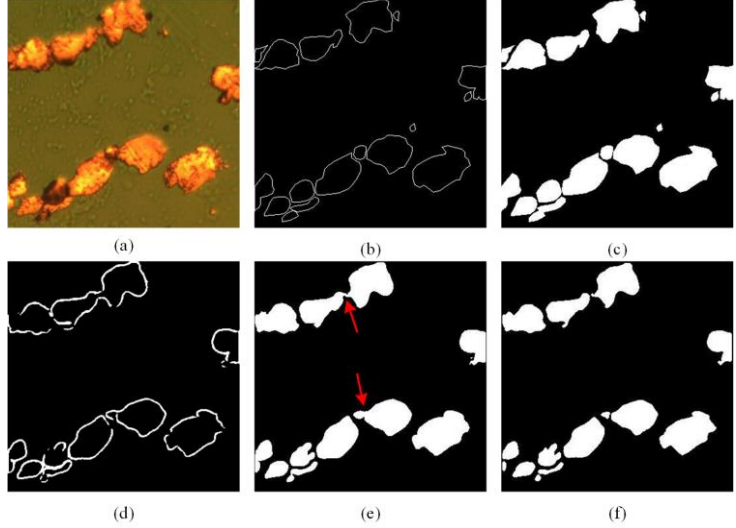

Figure 7. Morphology post-processing

In order to achieve the separation of overlapping debris, the wear particle region predicted by the network is subtracted from the wear particle edge predicted by the network. But it may cause some detail noise at this time. In order to further remove these detail noises, this paper adopts morphological open operation to remove isolated noise points and get the final segmentation result image. Figure 7 shows the results of this process.

## 3 Experiments

The experimental samples are obtained from the failed RV Reducer, and ferrograph images are then obtained. A total of 123 ferrograph images are collected in this experiment, of which 83 are used for model training and the remaining 40 are used for model testing. The experimental parameters are set as follows: the optimizer adopts Adam optimization method, the initial learning rate is set to 0.001, and the number of training epochs is set to 400. In addition, in order to prevent the model from overfitting, the online data augmentation is adopted. The methods of sample augmentation include random flipping, random rotation, random adding noise and random contrast change.

*3.1 Comparison of different models*

In order to compare the effects of existing models and the proposed models in this paper, experiments are carried out on the ferrograph image data set. The training batch of each iteration is set to 2, and the loss function adopts the square dice loss function. Except for the different model structure, other parameter settings are the same. All models adopt the same data augmentation method. The experimental results are shown in Table 1. In this study, dice per case [40] is used as the evaluation index of segmentation results. We add BN normalization to the original U-Net network, which is only used to segment the wear particle region but not used for edge detection. The OWSNet model without feature refine only adds the edge detection network to the improved U-Net model.

Table 1 Different model

| Model | Loss function | Boundary | Particle |
| --- | --- | --- | --- |
| U_Net | Square dice | × | 0.9009 |
| OWSNet-without-refine | Square dice | 0.2818 | 0.9165 |
| OWSNet -IN | Square dice | 0.2880 | 0.9194 |
| OWSNet-BN | Square dice | 0.2868 | 0.9156 |
| OWSNet-IN-BN | Square dice | 0.2863 | 0.9187 |
| OWSNet-BN-IN | Square dice | 0.2899 | 0.9196 |

## 3.2 Comparison of different loss functions

In the edge segmentation network of wear particles, there is a serious sample imbalance problem. Therefore, this paper proposes a square dice loss function to optimize the model parameters. Hence, the model can extract more accurate edges. Table 2 shows the segmentation results of the OWSNet in BN model with different loss functions. It can be seen from Table 2 that when the CE (cross-entropy) loss function is adopted, the result of edge segmentation is the worst. This is because the weight of each pixel loss function is treated equally by the CE loss function, while the number of wear debris edge pixels is far less than the number of background pixels in the image. Therefore, the prediction of all the background regions by the model will not produce too high loss, thus obtaining poor edge segmentation. When dice loss function or exponential logarithmic loss function is used, the edge segmentation is much better than that of CE loss function. In addition, compared with dice loss function and exponential logarithm loss function, the square dice loss function and exponential square dice loss function proposed in this paper also improves the result of edge segmentation.

Table 2 Different loss

| Model | Loss function | Boundary | Particle |
| --- | --- | --- | --- |
| OWSNet-IN-BN | CE | 0.0212 | 0.9190 |
| OWSNet-IN-BN | Dice | 0.2225 | 0.9181 |
| OWSNet-IN-BN | Exponential logarithmic dice | 0.2190 | 0.9218 |
| OWSNet-IN-BN | Square dice | 0.2863 | 0.9187 |
| OWSNet-IN-BN | Exponential square dice | 0.2929 | 0.9201 |

In order to further verify the effect of the loss function proposed in this paper, the following combinatorial loss function experiments are carried out, that is, the traditional CE loss function is adopted for the wear particle region segmentation network, and other loss functions are used for the wear particle edge segmentation network. The experimental results are shown in Table 3. It can be seen from Table 3 that the proposed square dice loss function achieves better edge segmentation.

Table 3 Different combinatorial loss

| Model | Loss function | Boundary | Particle |
| --- | --- | --- | --- |
| OWSNet-IN-BN | CE & CE | 0.0212 | 0.9190 |
| OWSNet-IN-BN | CE & Dice | 0.2214 | 0.9168 |
| OWSNet-IN-BN | CE & Exponential logarithmic dice | 0.2282 | 0.9203 |
| OWSNet-IN-BN | CE & Square dice | 0.2939 | 0.9205 |

## 3.3 Effects of different batch size

Due to the limitation of experimental conditions, the batch of model training cannot be set too large. The results show that when the training batch is small, different normalization operation modes have a great influence on the model results, especially batch normalization. BN has little effect on the model improvement when the training batch is small, while IN can achieve a better result. However, when the batch size increases, the performance of BN is improved. When the batch size reaches a certain upper limit, BN is better than IN. In addition, reference [37] shows that the combination of IN and BN can make the model solve the problem of image segmentation in different domains.

In order to explore the influence of different normalization on the results of the model, different training batches are set up and different models are tested. The experimental results are shown in Table 4. It can be seen from table 4 that BN is not suitable for the network when the training batch is small. Good results can be achieved by IN or combination of IN and BN

Table 4 Different batch size

| Model | Batch size | Boundary | Particle |
| --- | --- | --- | --- |
| OWSNet-IN-BN | 1 | 0.2815 | 0.9152 |
| OWSNet-IN-BN | 2 | 0.2863 | 0.9187 |
| OWSNet-IN-BN | 4 | 0.2907 | 0.9190 |
| OWSNet-IN-BN | 6 | 0.2889 | 0.9198 |
| OWSNet-BN-IN | 1 | 0.2790 | 0.9140 |
| OWSNet-BN-IN | 2 | 0.2899 | 0.9196 |
| OWSNet-BN-IN | 4 | 0.2930 | 0.9198 |
| OWSNet-BN-IN | 6 | 0.2956 | 0.9219 |
| OWSNet-IN | 1 | 0.2887 | 0.9187 |
| OWSNet-IN | 2 | 0.2880 | 0.9194 |
| OWSNet-IN | 4 | 0.2867 | 0.9174 |
| OWSNet-IN | 6 | 0.2956 | 0.9208 |
| OWSNet-BN | 1 | 0.2671 | 0.9018 |
| OWSNet-BN | 2 | 0.2868 | 0.9156 |
| OWSNet-BN | 4 | 0.2901 | 0.9221 |
| OWSNet-BN | 6 | 0.2898 | 0.9202 |

*3.4 Visualization*

In this paper, a new overlapping debris segmentation model is proposed, and a square dice loss function is proposed to overcome the sample imbalance problem. Section 3.1 to Section 3.3 of this paper makes quantitative comparison of different models and different loss functions. In order to further compare the segmentation effects of different models and different loss functions, this section will show the segmentation results of various methods. Figure 8 and Figure 9 show the segmentation results of different methods. As shown in Figure 8 and Figure 9, the ordinary U-Net network cannot achieve the segmentation of overlapping debris. Besides, the CE loss function cannot accurately segment the wear particle edge. However, dice loss function and exponential logarithmic loss function have some open edges. Although the square dice loss function and exponential square dice loss function also have some open edges, their edge segmentation results are better than other loss functions. Moreover, we also present the results of conventional wear particle segmentation methods in Figure 10. The traditional algorithms include k-means [9], watershed [41] and improved watershed methods [13]. In the k-means method, we set the number of cluster centres to 2 and 3 respectively. When the cluster is set to 3, we merge the results of the two wear particle region clusters to obtain the final output. Comparing the segmentation results of traditional algorithms and deep learning algorithms, it can be seen that the results of the deep learning algorithms are better than that of the traditional segmentation algorithms.

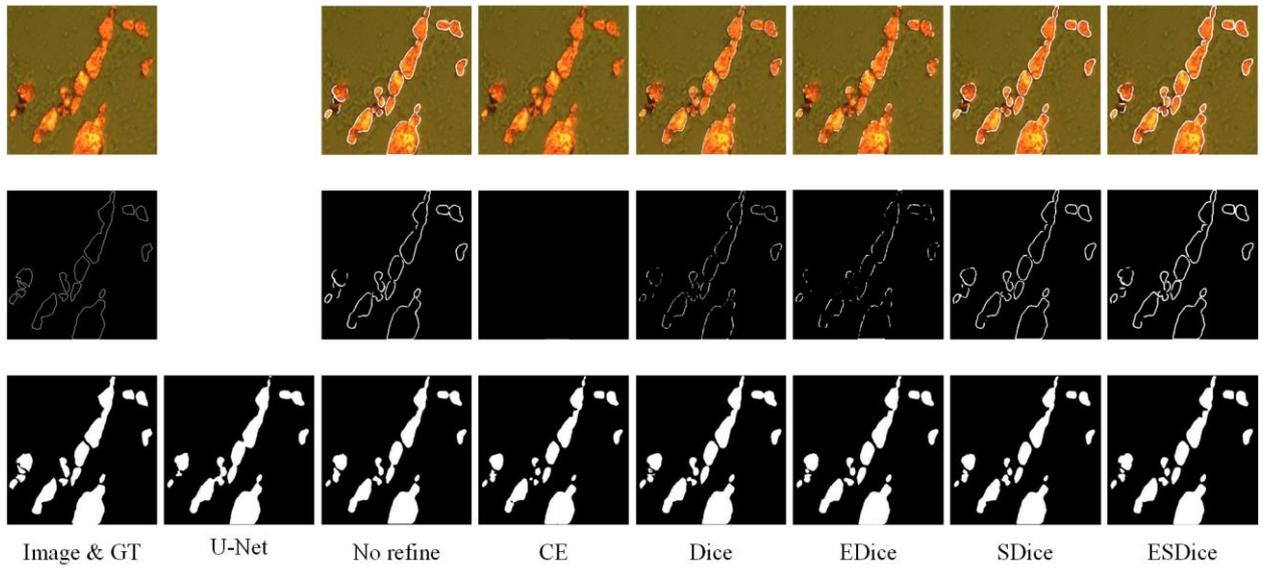

Figure 8. An example of segmentation result.

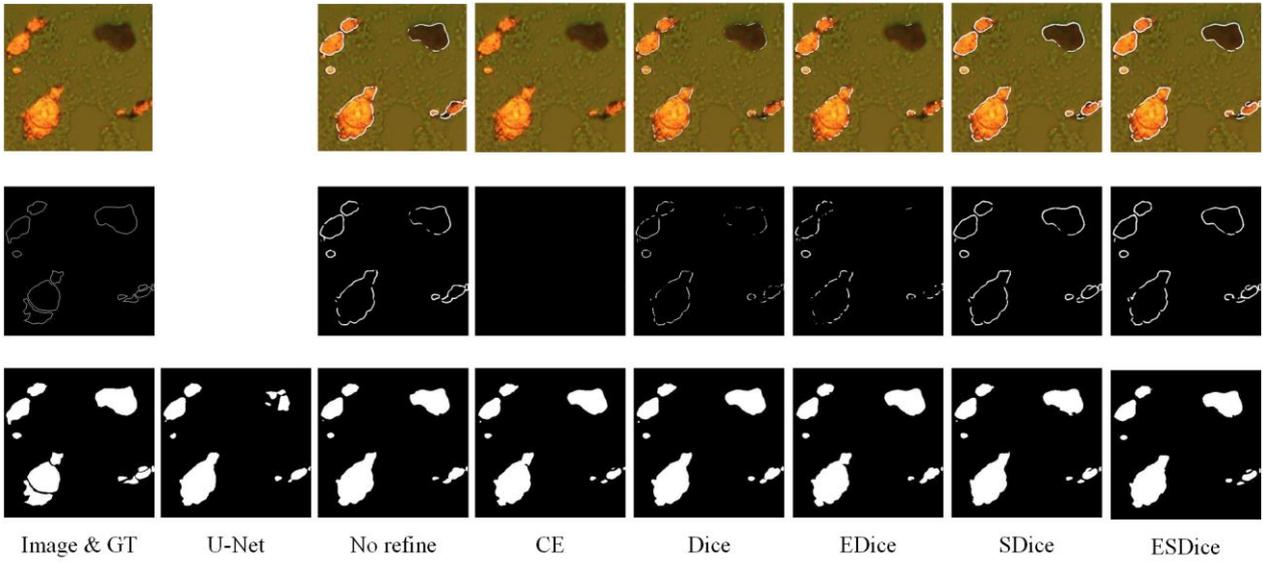

Figure 9. An example of segmentation result.

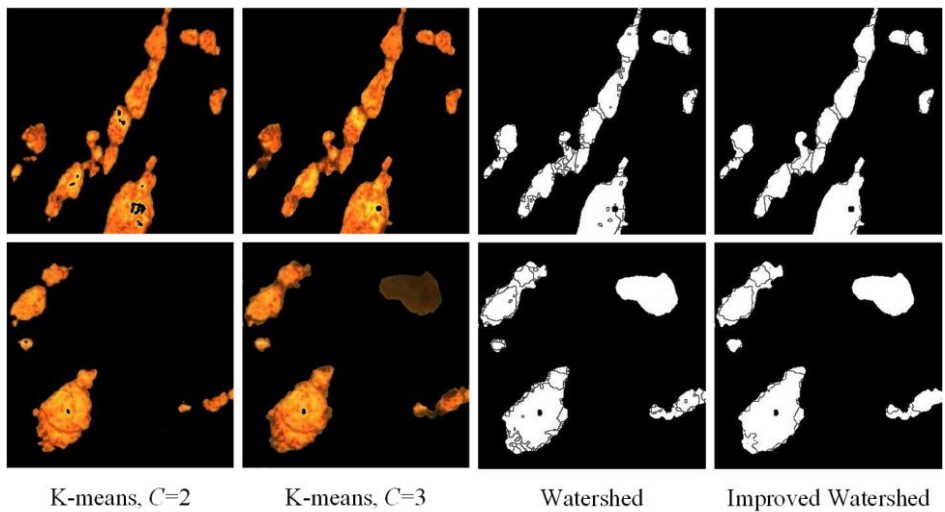

Figure 10. Segmentation results of conventional methods.

# 4 Conclusion

Aiming at the problems of few samples and overlapping debris, a new method of wear particle segmentation is proposed in this paper. Based on the U-Net model, the proposed method adds the branch of wear particle edge segmentation network, normalization and feature refine module. The experimental results show that the improved network can segment the debris region and the wear particle edge, and the segmentation of overlapping debris can be obtained by fusing the results of wear particle region and wear particle edge. In addition, in order to solve the problem of sample imbalance in particle edge segmentation, a new square dice loss function and an exponential square dice loss function are proposed. Experimental results show that the two loss functions can improve the segmentation results of wear particle edge.

# References

[1]. Roylance, B.J., Ferrography—then and now. Tribology international, 2005. 38(10): p. 857-862.
[2]. Myshkin, N.K., et al., Classification of wear debris using a neural network. Wear, 1997. 203-204(3): p. 658-662.
[3]. Cao, W., et al., Correction strategies of debris concentration for engine wear monitoring via online visual ferrograph. Proceedings of the Institution of Mechanical Engineers. Part J, Journal of engineering tribology, 2015. 229(11): p. 1319-1329.
[4]. Kumar, A. and S.K. Ghosh, Size distribution analysis of wear debris generated in HEMM engine oil for reliability assessment: A statistical approach. Measurement, 2019. 131: p. 412-418.
[5]. Wang, J., et al., A wear particle identification method by combining principal component analysis and grey relational analysis. Wear, 2013. 304(1-2): p. 96-102.
[6]. Wu, H., et al., Restoration of defocused ferrograph images using a large kernel convolutional neural network. Wear, 2019. 426-427: p. 1740-1747.
[7]. Peng, Y., et al., Oxidation wear monitoring based on the color extraction of on-line wear debris. Wear, 2015. 332-333: p. 1151-1157.
[8]. Yu, S., et al., Wear particle image segmentation method based on the recognition of background color. Tribology, 2007(05): p. 467-471.
[9]. Wang, J., et al., Ferrographic image segmentation by the method combining k-menas clustering and watershed algorithm. Journal of China university of mining & technology, 2013. 42(05): p. 866-872.
[10]. Wang, J., et al., The segmentation of wear particles in ferrograph images based on an improved ant colony algorithm. Wear, 2014. 311(1-2): p. 123-129.
[11]. Wang, J., et al., A Hybrid Method for the Segmentation of a Ferrograph Image Using Marker-Controlled Watershed and Grey Clustering. Tribology transactions, 2016. 59(3): p. 513-521.
[12]. Wu, T., et al., Imaged wear debris separation for on-line monitoring using gray level and integrated morphological features. Wear, 2014. 316(1-2): p. 19-29.
[13]. Peng, P., et al., Analysis of oxide wear debris using ferrography image segmentation. Industrial Lubrication and Tribology, 2019. 71(7): p. 901-906.
[14]. Isensee F., et al., nnu-net: Breaking the spell on successful medical image segmentation. arXiv preprint arXiv:1904.08128, 2019, 1: 1-8.
[15]. Chen, L., et al., DeepLab: Semantic Image Segmentation with Deep Convolutional Nets, Atrous Convolution, and Fully Connected CRFs. IEEE Transactions on Pattern Analysis and Machine Intelligence, 2018. 40(4): p. 834-848.
[16]. Shelhamer E., et al., Fully convolutional networks for semantic segmentation [J]. IEEE transactions on pattern analysis and machine intelligence, 2017, 39(4): 640-651.

[17]. Peng, P., et al., Ferrograph image classification. arXiv preprint arXiv:2010.06777, 2020.

[18]. Peng, P., et al., Wear particle classification considering particle overlapping. Wear, 2019. 422-423: p. 119-127.

[19]. Peng, Y., et al., A hybrid convolutional neural network for intelligent wear particle classification. Tribology international, 2019. 138: p. 166-173.

[20]. Peng, Y., et al., WP-DRnet: A novel wear particle detection and recognition network for automatic ferrograph image analysis. Tribology International, 2020. 151: p. 106379.

[21]. Wang, S., et al., Optimized CNN model for identifying similar 3D wear particles in few samples. Wear, 2020. 460-461: p. 203477.

[22]. Zhang, T., et al., CDCNN: A Model Based on Class Center Vectors and Distance Comparison for Wear Particle Recognition. IEEE Access, 2020. 8: p. 113262-113270.

[23]. Ronneberger O., et al., U-net: Convolutional networks for biomedical image segmentation[C]. International Conference on Medical image computing and computer-assisted intervention. Springer, Cham, 2015: 234-241.

[24]. Zhao, A., et al., Data augmentation using learned transformations for one-shot medical image segmentation [C]. Proceedings of the IEEE International Conference on Computer Vision. 2019: 8543-8553.

[25]. Shaban, A., et al., One-Shot Learning for Semantic Segmentation. arXiv preprint arXiv: 1709.03410, 2017.

[26]. Wang, K., et al., PANet: Few-Shot Image Semantic Segmentation with Prototype Alignment [C]. Proceedings of the IEEE International Conference on Computer Vision. 2019: 9197-9206.

[27]. Zhang, C., et al., CANet: Class-Agnostic Segmentation Networks with Iterative Refinement and Attentive Few-Shot Learning [C]. Proceedings of the IEEE International Conference on Computer Vision. 2019: 5217-5226.

[28]. He, Kaiming., et al., Mask r-cnn [C]. Proceedings of the IEEE International Conference on Computer Vision. 2017: 2961-2969.

[29]. Chen, H., et al., BlendMask: Top-Down Meets Bottom-Up for Instance Segmentation [C]. Proceedings of the IEEE International Conference on Computer Vision. 2020:8573-8581.

[30]. Takikawa T., et al., Gated-scnn: Gated shape cnns for semantic segmentation[C]. Proceedings of the IEEE International Conference on Computer Vision. 2019: 5229-5238.

[31]. Acuna D., et al., Devil is in the edges: Learning semantic boundaries from noisy annotations [C]. Proceedings of the IEEE Conference on Computer Vision and Pattern Recognition. 2019: 11075-11083.

[32]. Fu J., et al., Dual attention network for scene segmentation[C]. Proceedings of the IEEE Conference on Computer Vision and Pattern Recognition. 2019: 3146-3154.

[33]. Ioffe S., et al., Batch normalization: Accelerating deep network training by reducing internal covariate shift. arXiv preprint arXiv:1502.03167, 2015.

[34]. Singh S., et al., Filter response normalization layer: Eliminating batch dependence in the training of deep neural networks [C]. Proceedings of the IEEE/CVF Conference on Computer Vision and Pattern Recognition. 2020: 11237-11246.

[35]. Ioffe, S., Batch Renormalization: Towards Reducing Minibatch Dependence in Batch-Normalized Models [C]. Advances in neural information processing systems. 2017: 1945-1953.

[36]. Ulyanov D., et al., Instance normalization: The missing ingredient for fast stylization. arXiv preprint arXiv:1607.08022, 2016.

[37]. Pan X., et al., Two at once: Enhancing learning and generalization capacities via ibn-net [C]. Proceedings of the European Conference on Computer Vision (ECCV). 2018: 464-479.

[38]. Milletari F., et al., V-net: Fully convolutional neural networks for volumetric medical image segmentation [C]. 2016 fourth international conference on 3D vision (3DV). IEEE, 2016: 565-571.

[39]. Wong K C L., et al.,3D segmentation with exponential logarithmic loss for highly unbalanced object sizes [C]. International Conference on Medical Image Computing and Computer-Assisted Intervention. Springer, Cham, 2018: 612-619.

[40]. Bilic, P., et al., The Liver Tumor Segmentation Benchmark (LiTS). arXiv preprint arXiv:1901.04056, 2019.

[41]. Vincent, L., et al., Watersheds in digital spaces: an efficient algorithm based on immersion simulations. IEEE transactions on pattern analysis and machine intelligence, 1991. 13(6): p. 583-598.